**ARTICLE**

# INDIC-DIALECT: A Multi-Task Benchmark to Evaluate and Translate in Indian Language Dialects

Tarun Sharma,*† Manikandan Ravikiran,† Sourava Kumar Behera,† Pramit Bhattacharya,‡ Arnab Bhattacharya,‡ and Rohit Saluja†

†Indian Institute Of Technology, Mandi, 175075, HP, India
‡Indian Institute Of Technology, Kanpur, India
*Corresponding author. Email: tarunsharma7845@gmail.com

**Abstract**

Recent NLP advances focus primarily on standardized languages, leaving most low-resource dialects under-served especially in Indian scenarios. In India, the issue is particularly important: despite Hindi being the third most spoken language globally (over 600 million speakers), its numerous dialects remain underrepresented. The situation is similar for Odia, which has around 45 million speakers. While some datasets exist which contain standard Hindi and Odia languages, their regional dialects have almost no web presence. We introduce INDIC-DIALECT, a human-curated parallel corpus of 13k sentence pairs spanning 11 dialects and 2 languages: Hindi and Odia. Using this corpus, we construct a multi-task benchmark with three tasks: dialect classification, multiple-choice question (MCQ) answering, and machine translation (MT). Our experiments show that LLMs like GPT-4o and Gemini 2.5 perform poorly on the classification task. While fine-tuned transformer based models pretrained on Indian languages substantially improve performance e.g., improving F1 from 19.6% to 89.8% on dialect classification. For classification and translation related MCQ tasks, we observe that state specific models perform better than combined model trained on all dialects. For MT, we try multiple approaches: i) rule-based: bilingual dictionary based translation, ii) AI-based: finetuning transformer model for translation in usual way, iii) rule-based followed by AI, and iv) hybrid AI approach. For dialect to language translation, we find that hybrid AI model achieves highest BLEU score of 61.32 compared to the baseline score of 23.36. Interestingly, due to complexity in generating dialect sentences, we observe that for language to dialect translation the "rule-based followed by AI" approach achieves best BLEU score of 48.44 compared to the baseline score of 27.59. INDIC-DIALECT thus is a new benchmark for dialect-aware Indic NLP, and we plan to release it as open source to support further work on low-resource Indian dialects.

## 1. Introduction

There has been a substantial progress in multilingual NLP recently, driven by multiple artificial intelligence (AI) based models and large-scale benchmarks spanning several languages, though the languages are unevenly distributed (Joshi et al. 2020). Most multilingual models are trained and evaluated on standardized languages, often derived from newswire, wikipedia, and other formal/semi-formal sources (Kreutzer et al. 2022). In contrast, regional dialects that dominate spoken interaction and informal communication remain severely underrepresented, especially in India. As a result, the Indic models that are "multilingual" in a formal sense fail on the dialects that a significant proportion of population actually use in daily life (see Fig. 1). In India, the issue is prominent due to some of the important dialects having limited or almost no presence on the web. The country officially recognizes 22 scheduled languages and reports 121 languages and over 19,000 dialects in active use (Census of India 2011: Language Data). While languages such as Hindi and Odia serve as standardized written and administrative forms, they coexist with a rich ecosystem of regional dialects that differ systematically in phonology, lexicon, and syntax. Though there are some works on dialect identification in speech (Indian Institute of Science, ARTPARK, and Google 2022), many of these dialects are underrepresented in mainstream NLP resources. From the modeling perspective, dialects are not merely "noisy" variants of standard languages. Instead, they often encode systematic and productive transformations—such as regular phonetic shifts, distinct inflectional paradigms, or region-specific lexical inventories-that cannot be captured by simply treating dialectal text as noisy variant of the corresponding language. This raises two intertwined challenges. First, there is a resource challenge: the absence of curated corpora that represent dialectal usage in a way suitable for modeling and evaluation. Second, there is a methodological challenge: the need to understand whether existing architectures and training strategies are adequate for dialect processing, or whether specialized approaches are required. In this paper, we investigate these challenges in the context of Hindi and Odia dialects. We introduce INDIC-DIALECT, a parallel, multi-dialect corpus that systematically pairs dialectal sentences with their standardized Hindi and Odia counterparts. The corpus comprises 13,000 manually annotated sentence pairs, covering 2 languages and 11 dialects that are selected based on factors such as geographical spread, sociolinguistic relevance, and minimal existing web presence. The INDIC-DIALECT dataset cover multiple dialects from three states of India: Himachal Pradesh, Uttar Pradesh, and Odisha. All data is produced and validated by native speakers to ensure linguistic fidelity and dialectal authenticity. Also, INDIC-DIALECT is not limited to serving as a static parallel corpus. We leverage it to build a multi-task benchmark that reflects complementary aspects of dialectal processing:

1. Dialect classification, where the goal is to classify a given sentence into one of the supported dialects, probing the ability of models to encode dialect-sensitive lexical and orthographic cues.

2. Multiple-choice question (MCQ) answering, which is a task designed to check the translation understanding of the models without actually translating. The MCQ dataset contain i) a sentence in dialect as a question, ii) correct translation of (i) to the corresponding language as one of the option, and iii) the three sentences, close to (ii), in corresponding language as the distractors.

3. Machine translation (MT) between dialects and their corresponding languages, treated in two directions: language to dialect translation and dialect to language translation.

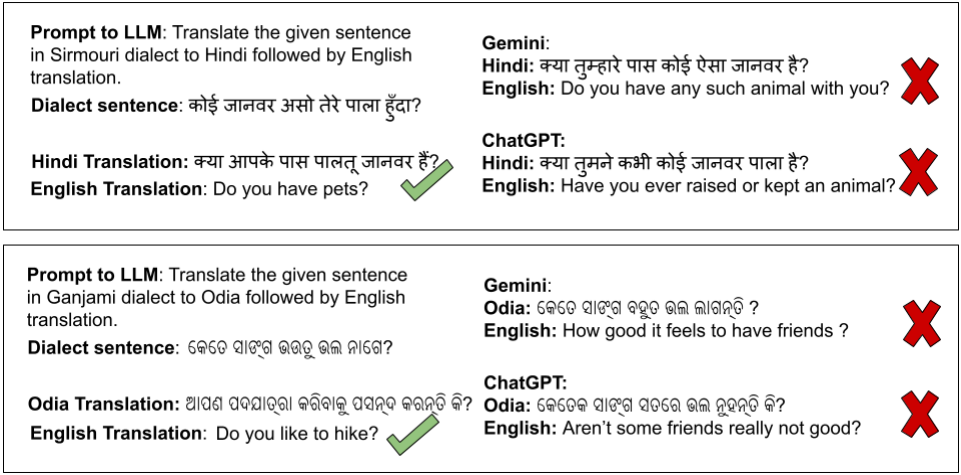

**Figure 1.** Top-Left: A promt to Large Language Models (LLM) to translate a sentence in Sirmouri dialect (from Himachal Pradesh) to Hindi followed by English. The correct Hindi and English Translations are also shown only for reference. Top-Right: Incorrect translations from Gemini and ChatGPT, showcasing that LLMs are unable to handle dailect translations to Hindi (English translations are shown only for reference). Bottom: Another example for Ganjami dialect (from Odisha) to Odia translation by LLMs.

Our initial experiments on the INDIC-DIALECT dataset provide interesting findings. First, we observe that fine-tuned Indian language models (e.g., IndicBERT) substantially improve classification performance by 70% compared to multilingual Large Language Models (LLMs) based zero-shot baselines like GPT-4o and Gemini. Specifically, the fine-tuning raises F1 score from 19.6 to 89.8, revealing that general-purpose LLMs, despite their broad coverage, do not automatically handle under-represented dialectal forms. The findings reveal that targeted fine-tuning on modest amounts of dialectal data can yield disproportionate gains in downstream performance compared to relying on generic zero-shot capabilities. Additional experiments on classification and translation based MCQ tasks reveal that state specific models work better than a collective model trained on all the dialects. Furthermore, To benchmark Machine Translation (MT) performance on Indic dataset, we investigate four distinct methodologies ranging from traditional linguistic techniques to modern AI approaches: (i) a Dictionary Based method which utilizes bilingual dictionaries to translate given source sentence to target sentence by simply replacing words; (ii) an AI-based approach in which we fine-tune transformer model on our INDIC-DIALECT dataset and they act as baseline models in our MT experiments. (iii) a Rule-based approach followed by AI; in this approach we are providing bilingual dictionary based generated sentence as input to the AI model. (iv) a Hybrid AI-based strategy: In this approach we provide a concatenation of "bilingual dictionary based translation" and "source sentence" as input to AI model for translation (see Fig. 4). Our experiments reveal that the optimal strategy depends on the translation direction. For Dialect to Standard Language, where the goal is normalization, the Hybrid + AI model yields the best performance, achieving a BLEU score of 61.3 compared to the baseline of 46.5. Conversely, the Standard Language to Dialect direction requires generating complex, region-specific morphology. In this more challenging setting, the Rule + AI-based approach proves superior, effectively guiding the generation process to improve the BLEU score from 27.6 to 48.4. The key contributions of our work are:

1. We introduce INDIC-DIALECT, the first parallel, multi-dialect corpus for Hindi and Odia, comprising 13,000 sentence pairs across 11 Indian dialects from three states, curated and validated by native speakers. To the best of our knowledge, INDIC-DIALECT is a parallel corpus covering maximum number of Indian dialects currently available as open source.

2. We construct a multi-task benchmark for dialect identification, MCQ, and MT, and provide a comprehensive evaluation comparing fine-tuned Indic encoders and strong zero-shot LLMs.

By releasing INDIC-DIALECT, related tasks, and the associated baselines, we aim to enable a more realistic and inclusive evaluation of NLP systems in Indic settings.

## 2.   Related Work

The success of Large Language Models (LLMs) on high-resource languages has created a significant research push to extend these capabilities to the world's diverse linguistic landscape (Hu et al. 2020). In this section, we position our work with respect to prior research on (i) benchmarks and resources for standard Indian languages, (ii) emerging datasets for dialects and language varieties, and (iii) methods for low-resource and dialectal machine translation.

### 2.1   Benchmarks for Standard Indian Languages

A growing body of work has developed benchmarks and resources for standard Indian languages, primarily focusing on Hindi and a subset of other scheduled languages. Early efforts centered on monolingual tasks such as part-of-speech tagging, dependency parsing, and named entity recognition, often using newswire or curated text. Subsequently, broader benchmark suites were proposed to support evaluation across multiple tasks. For example, benchmarks in the spirit of IndicGLUE (Doddapaneni et al. 2023 ) provide collections of classification, NLI, and question answering (QA) tasks over standardized Hindi and several other Indian languages, while datasets like Naamapadam (Mhaske et al. 2022) and Indic-NLG (Kumar et al. 2022) target named entity recognition and natural language generation in standard scripts. More recently, multilingual reading comprehension and QA benchmarks (e.g., cross-lingual datasets analogous to Belebele (Bandarkar et al. 2023)) have begun to include standard Hindi as one of many target languages.

Despite these advances, existing benchmarks share two important limitations from a dialectal perspective. First, they operate almost exclusively on standardized forms of language, with orthography and morphology that closely follow official or educational norms. Second, they are designed as single-variety benchmarks: they do not contain parallel examples that link dialectal and standard forms, nor do they include tasks that explicitly probe dialectal variation. As a result, while these resources have been crucial in driving progress for standard Hindi and related languages, they offer no direct support for evaluating how models handle dialects, nor do they provide supervision suitable for training dialect-aware systems.

### 2.2   Resources for Dialects and Language Varieties

Outside the Indic context, several lines of work have investigated dialect identification, language variety classification, and regional language modeling. In Arabic, for instance, shared tasks and benchmarks have focused on distinguishing Modern Standard Arabic from multiple regional dialects (Bouamor, Hassan, and Habash 2019) (Abdul-Mageed et al. 2020), often based on social media text. Similar efforts exist for Chinese language varieties (Xu, Wang, and Li 2017), Swiss German (Scherrer and Rambow 2010), and other language families where clear dialect continua are present. These datasets typically frame dialect processing as a classification problem, asking models to assign a dialect label to each input.

In India, dialect resources remain comparatively sparse. Some corpora compiled by initiatives such as national language documentation centers and linguistics departments provide speech recordings for a subset of dialects, often to support phonetic analysis or dialect atlas projects. These collections, while valuable for understanding sound patterns and pronunciation, rarely include aligned text or parallel standard–dialect pairs, making them difficult to use for downstream NLP tasks such as translation or textual QA.

A smaller number of datasets target textual dialect identification within a single language family (for example, dataset by the VarDial 2018 (Zampieri et al. 2018), HinDialect dataset (Bafna 2022)). These resources have generally been constructed by collecting user-generated content from social media and labeling it by dialect or region. Another popular avenue in direction of dialect is speech technology, several corpora which contain implicit dialectal information has recently released. Such as the Bangla Dialecto (Samin et al. 2024) project, has demonstrated the feasibility of creating parallel corpora for dialect-to-standard language translation with its focus on the Noakhali dialect of Bengali, This work includes constructing a 10 hours diverse dataset with dialectal speech and then translating the dialect text to standard Bangla text. Similarly, the LDC-IL's Tamil and Punjabi speech corpora collected by (Choudhary 2021) from distinct dialect regions to improve Automatic Speech Recognition (ASR) robustness. More targeted efforts, such as AI4Bharat's Lahaja by (Javed et al. 2024) benchmark, provide explicit accent-specific data for Hindi ASR evaluation.

**Table 1.** Dialects covered in the INDIC-DIALECT dataset, grouped by states they belong to.

| States | Dialects Included |
|---|---|
| Himachal Pradesh | Kulluvi, Bilaspuri, Mandyali, Sirmouri |
| Uttar Pradesh | Meerut, Bhatner, Muzaffarnagar |
| Odisha | Sambalpuri, Ganjami, Baleswari, Desia |

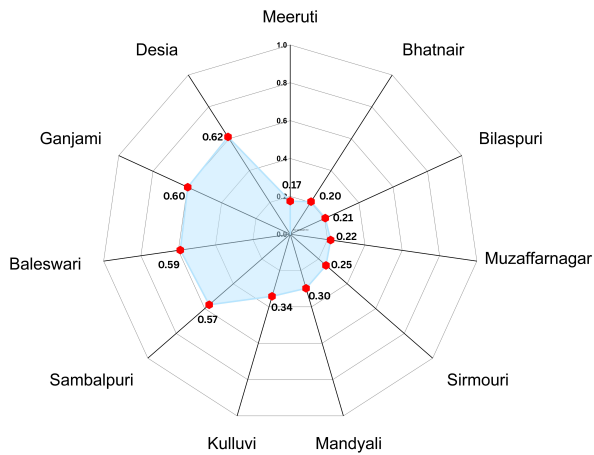

**Figure 2.** Graphical representation of different dialects based on lexical distance from Hindi language which is most prominent language among Indian population. Note that only in the shown graphical representation: the sentences from Odisha dialects are transliterated to Devanagari (script used for Hindi) to appropriately find lexical overlaps with Hindi.

While above mentioned datasets are useful for training classifiers, they exhibit several limitations relative to our goals: they (i) are usually monolingual and non-parallel, (ii) focus on a single task (dialect identification), and (iii) do not support-task evaluation involving translation. To the best of our knowledge, there is no existing Indic resource that simultaneously offers: (a) manually curated, parallel dialect–standard sentence pairs, (b) coverage of multiple dialects across at least two language families (here, Hindi and Odia), and (c) a multi-task design enabling dialect identification, MCQ, and translation within a single benchmark. The INDIC-DIALECT is designed carefully to fill the above-mentioned gap.

## 3.    INDIC-DIALECT

To enable systematic research on low-resource Indian dialects, we introduce INDIC-DIALECT, a novel benchmark comprising a multi-way parallel corpus and a suite of evaluation tasks. This section details the corpus curation methodology, data quality validation, and the formulation of the experimental tasks.

**Table 2.** An example from the proposed INDIC-DIALECT parallel corpus showing translations from standard Hindi/Odia into various regional dialects. The example illustrates the lexical and syntactic variations captured in the dataset.

(English translation and roman transliterations in parenthesis are shown only for better understanding.)

| Standard Languages Hindi/ Odia and their dialectal counterparts in 11 dialects | |
|---|---|
| English: My teacher explained how cosmic rays are formed. | Kulluvi (Himachal Pradesh): म्हारे शिक्षक दसु कि ब्रह्मांडीय किरणें केन्डे बनु सा । (mhārē śikṣaka dasu ki brahmāmḍīya kiraṇēm kēṇḍe banu sā .) |
| Hindi: मेरे शिक्षक ने बताया कि ब्रह्मांडीय किरणें कैसे बनती हैं। (mērē śikṣaka nē batāyā ki brahmāmḍīya kiraṇēm kaisē banatī haim.) | Mandyali (Himachal Pradesh): मेरी शिक्षके दसया कि ब्रह्माडीय किरणा कियां बनां ई (mērī śikṣakē dasayā ki brahmāḍīya kiraṇa kiyām banām ī) |
| Odia: ମୋ ଶିକ୍ଷକ ବ୍ୟାଖ୍ୟା କରିଥିଲେ ଯେ କିପରି ମହାଜାଗତିକ ରଶ୍ମି ସୃଷ୍ଟି ହୁଏ। (mō śikṣaka byākhyā karithilē yē kipari mahājāgatika raśmi sṛṣṭi huē.) | Sirmouri (Himachal Pradesh): मेरे शिक्षकों ने बताया कि ब्रह्मांडीय किरणे किशी बनो। (mērē śikṣakōm nē batāyā ki brahmāmḍīya kiraṇē kiśī banō .) |
| Ganjami(Odisha): ବିଜ୍ଞାନିକମାନେ ଜଲବାୟୁ ପରିବର୍ତ୍ତନର ପ୍ରଭାବ ଉପରେ ଗୋଟେ ରିପୋର୍ଟ ପ୍ରକାଶ କରିଛନ୍ତି। (bijñānikamānē jalabāyu paribarttanara prabhāba uparē gōṭe ripōrṭa prakāśa karichanti.) | Bilaspuri (Himachal Pradesh): मेरे शिक्षक ऐ बताया कि ब्रह्मांडीय किरणें किया बना ई। (mērē śikṣaka ai batāyā ki brahmāmḍīya kiraṇēm kiyā banā ī.) |
| Desia (Odisha): ମୋର ପାଟପଢ଼ାଉ ବ୍ୟାଖ୍ୟା କରି ରେଲାଏ ଯେ କେନ୍ତା ମହାଜାଗତିକ ତରାସ ସୃଷ୍ଟି ହେସି (mōr pāṭapaḍhāu byākhyā kari rēlāē yē kēntā mahājāgatika tarās sṛṣṭi hēsi.) | Bhatnair (Uttar Pradesh): मेरे मौस्टर नै बतायौ कि ब्रह्मांडीय किरणे कैसे बने है। (mērē maustara nai batāyau ki brahmāmḍīya kiraṇe kaisai banai hai.) |
| Sambalpuri (Odisha): ମୋର ଗୁରୁ ବ୍ୟାଖ୍ୟା କରିଥିଲେ ଜେ କେନ୍ତା ମହାଜାଗତିକ ରଶ୍ମି ସୁରୁ ହେସି। (mōr guru byākhyā karithilē jē kēntā mahājāgatik rasmi suru hēsi .) | Meeruti (Uttar Pradesh): मेरे शिक्षक ने बताया कि ब्रह्मांडीय किरणें कैसे बनती हैं। (mērē śikṣaka nē batāyā ki brahmāmḍīya kiraṇēm kaisē banatī haim.) |
| Baleswari (Odisha): ମୋର ଶିକ୍ଷକ ବର୍ଣ୍ଣନା କରୁଥିଲେଯେ କେମିତି ମହାଜାଗତିକ ରଶ୍ମି ସୃଷ୍ଟି ହୁଏ। (mōr śikṣaka barṇṇanā karthilēyē kēmiti mahājāgatika raśmi sṛṣṭi huē .) | Muzaffarnagar (Uttar Pradesh): मेरे मास्टर ने बतायो कि ब्रह्मांड की किरण किस तरह स बना सैं। (mērē māstara nē batāyō ki brahmāmḍa kī kiraṇa kisa taraha sa banā saim.) |

### 3.1  Corpus Construction

The foundation of our benchmark is a human-curated parallel corpus comprising 13,000 sentence pairs. The corpus is hierarchically structured around two scheduled languages, Hindi and Odia, and 11 associated regional dialects. As shown in Table 1, the 11 dialects are spoken across three Indian states: Himachal Pradesh (Kulluvi, Bilaspuri, Mandyali, Sirmouri), Uttar Pradesh (Meerut, Bhatner, Muzaffarnagar), and Odisha (Sambalpuri, Ganjami, Baleswari, Desia). Unlike traditional corpora that focus solely on standard language to dialect pairs, our dataset is designed as an N-way parallel corpus as shown through an example in Table 2. The example shows a common sentence in Hindi, Odia, and English (English translation shown only for better understanding), and corresponding translations in the 11 dialects (with transliteration in bracket of each dialect translation shown only for illustration). Though the parallel structure enable tasks related to dialect-to-dialect translation, we focus on dialect-to-language and language-to-dialect translation tasks in this work. As shown in Fig. 2, the dialects are selected based on their varying degree of lexical proximity to Hindi, the most widely spoken language in India.

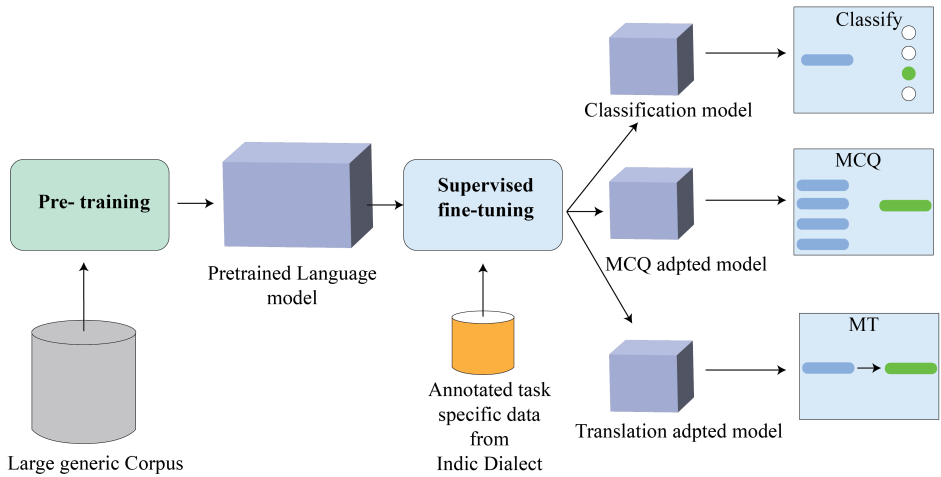

**Figure 3.** Finetuning the pre-trained transformer based model for three different tasks related to INDIC-DIALECT dataset.

### 3.2 Annotator Recruitment and Ethics

Following standard practices for linguistic data collection (NLLB Team 2022), the dialectal data is produced and curated entirely by native speakers (Nekoto, Kreutzer, Marivate, et al. 2020). We recruit 23 annotators (2 per dialect, and 1 for Odia) and 11 experts (1 per dialect) from the specific regions where these dialects are spoken. All annotators are compensated for their linguistic expertise at a rate exceeding the local minimum wage, ensuring ethical and fair labor practices throughout the data creation process. The 1000 Hindi sentences are initally created by experts covering different domains like science, history, culture, technology, arts, literature, language and linguistics, etc. The annotators help translate the Hindi sentences to regional dialects/Odia with expert validation as final step.

### 3.3 Quality Assurance and Validation

To ensure the reliability and linguistic fidelity of our human-curated dataset, we conducted a rigorous inter-annotator agreement (IAA) study. For the tasks requiring subjective judgment—namely MT (machine translation). A subset representing 5% of the total data is initially annotated independently by two native speakers for each dialect. We then calculated Cohen's Kappa (Cohen 1960) to measure the level of agreement while accounting for chance. The agreement is consistently high, achieving an average Kappa score of 0.89, reason for which is the annotation guidelines requiring native speakers to prioritize six core dimensions: naturalness of expression, orthographic consistency, local vocabulary selection, morphosyntactic adaptation, appropriate handling of loanwords, and preservation of sociolinguistic register. The high level of inter-rater reliability validates the clarity of our annotation guidelines and confirms that INDIC-DIALECT is a consistent and high-quality resource for model evaluation. The remaining 95% data is then translated by annotators, followed by the expert validation step.

### 3.4 Task Formulation

Using the INDIC-DIALECT corpus, we perform three distinct tasks to evaluate a range of models capabilities, from text generation to natural language understanding.

1. Dialect Classification: For the classification task, sentences from all dialects are compiled into a single dataset. Each sentence is paired with its corresponding dialect name, which serves as the ground-truth label for a multi-class classification problem.

2. Multiple-Choice Question (MCQ): To mimic how a human guesses the meaning of an unfamiliar dialect, our MCQ setup presents a single dialect sentence (the question) and four candidate interpretations in the standard language. Exactly one option is the correct translation taken from our parallel corpus; the other three are hard distractors created to be semantically very similar near-misses using approach given by Zellers et al. 2019. This design forces models to rely on dialect-specific cues—rather than superficial word overlap—when selecting the correct answer.

3. Machine Translation (MT): The parallel corpus is used directly to create the MT task. The data is formatted with source sentences in one language/dialect and target reference translations in the other language/dialect, supporting evaluation in both directions: language-to-dialect and dialect-to-language.

As shown in Fig. 3, we fine-tune different transformer based models, pre-trained on large corpus, for the tasks mentioned above (see Sec. 4 for more details on different models).

For machine translation, we examine four distinct methodologies, encompassing traditional linguistic techniques and contemporary AI approaches: (i) Rule-based: a method that uses bilingual word-level dictionaries (derived manually using training sets) to translate a source sentence into a target sentence simply by replacing words with their translations (see translation in orange in top-right of Fig. 9), (ii) an AI-based method that fine-tunes a transformer model on the INDIC-DIALECT dataset (excluding dictionary based translation in orange and corresponding tokens in orange from Fig. 9), (iii) a Rule-based approach followed by AI; in this case, we give the AI model a sentence that was made using a bilingual dictionary as input (excluding source sentence and corresponding tokens in blue from Fig. 9), and (iv) a hybrid AI stratergy: we give the AI model a combination of "bilingual dictionary based translation" and "source sentence" as input for translation in this method (as shown in Fig. 9).

## 4.   Experiments and Results

To evaluate the performance of models on our INDIC-DIALECT datset, we conducted a series of experiments across three tasks. This section details our experimental setup, presents the quantitative results, and provides a qualitative analysis of the key findings. For fine-tuning tasks, the INDIC-DIALECT corpus is splited using a 70:30 (train:test) ratio, To ensure stable and reliable results, each fine-tuning experiment was conducted over multiple trials (n=5) with different random seeds. The results presented in the following tables represent the average scores from these trials.

All experiments we conduct on compute node equipped with NVIDIA RTX A6000 (48GB) GPU. We implement our models using the PyTorch framework and the Hugging Face Transformers library. For optimization, we employed AdamW with a consistent learning rate of 2e-5 across all transformer-based models. A linear learning rate scheduler is used with a warm-up period of 10% of the total training steps to stabilize the training process. The fine-tuning is standardized with a global batch size of 32. To ensure robust convergence, the classification and MCQ models are trained for 100 epochs, while the more resource-intensive machine translation models are trained for 150 epochs. In all cases, the best-performing checkpoint are selected based on the lowest validation loss.

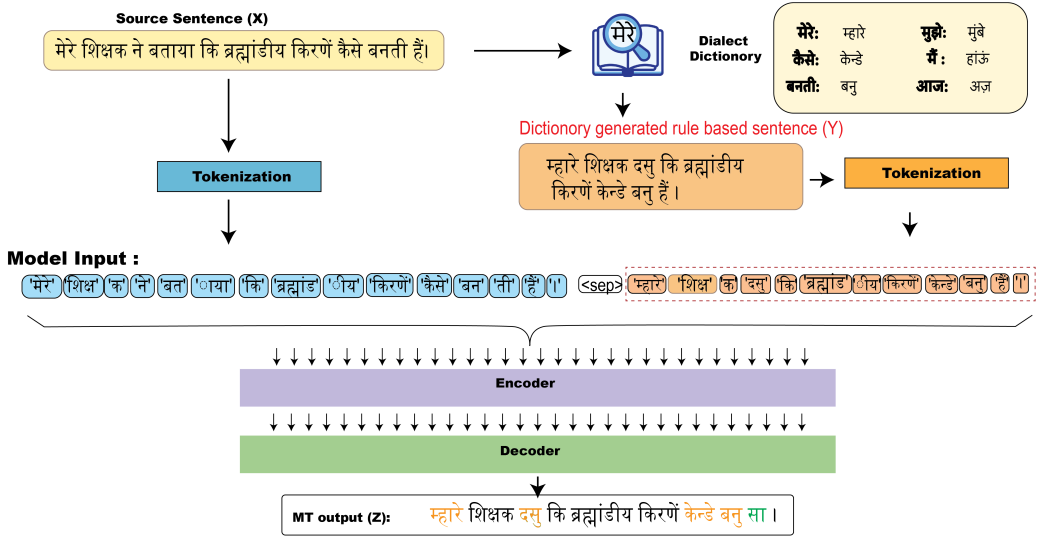

**Figure 4.** Hybrid approach for machine translation: bilingual word-level dictionary based translation concatenated with source sentence is model's input (if only the orange tokens from dictionary based sentence are given to the model than it act as "rule based followed by AI" approach).

**Table 3.** Dialect Classification Results for zero-shot (ZS) & fine-tuned (FT) models.

| Model | Prec. | Recall | F1 |
|---|---|---|---|
| ChatGPT-4o (ZS) | 3.93 | 6.25 | 4.25 |
| GEMINI 2.5 pro (ZS) | 19.32 | 17.82 | 19.62 |
| BART (FT) | 41.14 | 41.01 | 42.92 |
| Distilbert-base (FT) | 46.50 | 43.32 | 46.30 |
| mbart-large-50 (FT) | 72.98 | 72.60 | 72.51 |
| muril-base (FT) | 76.97 | 76.46 | 76.54 |
| **IndicBERT (FT)** | **89.70** | **89.60** | **89.75** |

## 4.1    Dialect Classification

For the dialect classification experiment, we evaluate the ability of various models to predict the correct dialect from a given sentence using Precision, Recall, and F1-score as metrics. We benchmark two large language models (LLMs) for zero-shot setting, ChatGPT-4o (OpenAI 2023) and GEMINI 2.5 pro (Reid, Savinov, Teplyashin, et al. 2024), against multiple fine-tuned models: BART (Lewis, Liu, Goyal, et al. 2020), Distilbert-base-multilingual-cased (Sanh et al. 2019), mbart- large (Liu et al. 2020), Muril (Khanuja et al. 2021) and IndicBERT V2. (Gala, Behl, et al. 2023) To get ChatGPT-4o and GEMINI 2.5 pro to classify text by language, prompt given is as: "Identify the dialect of the following text and respond with only the dialect name: <dialect text to classify>". Both zero-shot models perform poorly (3), with CHAT GPT-4o (F1-score of 04.25) failing to surpass a random baseline (score of 1/11 or 9.99) and GEMINI 2.5 pro achieving an F1-score of only 19.62. While all fine-tuned models outperform the zero-shot baselines, the key finding is the superior performance of the model pre-trained specifically on Indian languages. IndicBERT V2, which is pretrained on 24 Indian languages[1], emerge as the top-performing model, achieving the highest F1-score

---
1. https://ai4bharat.iitm.ac.in/areas/model/LLM/IndicBERTv2

**Table 4.** Performance comparison of the combined model with state-specific fine-tuned models for dialect classification.

| Model | Test Dataset | Prec. | Recall | F1 |
|---|---|---|---|---|
| All Dialect Combined Model | All Dialects | 89.70 | 89.60 | 89.75 |
| | Himachali Dialects | 91.75 | 89.05 | 91.5 |
| | Uttar Pradesh Dialects | 71.33 | 72.18 | 71.66 |
| | Odisha Dialects | 89.25 | 88.75 | 89.77 |
| Himachal | Himachali Dialects | 93.42 | 93.42 | 93.46 |
| Uttar Pradesh | Uttar Pradesh Dialects | 83.51 | 82.18 | 83.50 |
| Odisha | Odisha Dialects | 91.15 | 90.73 | 90.77 |

of 89.75.

The general multilingual models, Distilbert and BART, show only moderate success with F1-scores of 46.30 and 42.92, respectively. The mBART and MuRIL vastly outperform BART and DistilBART for Indian dialect tasks primarily due to vocabulary overlap, script familiarity, and pre-training objectives (mBART is pretrained on 4 Indian languages and several other languages, and MuRIL is pretrained on 17 Indian languages). While BART and Distilbertis are powerful models for English, they are fundamentally not-suited for Indian dialects due to their limited pretraining with Indian language data. The experiment demonstrates that while fine-tuning on in-domain data is crucial, leveraging a model pretrained on the specific language family provides a distinct advantage for the nuanced task of dialect identification. Given substantial performance advantage of IndicBert v2, we select IndicBERT V2 as the foundational model for all the subsequent experiments.

Based on the observation that the IndicBERT V2 model provide the strongest baseline, we hypothesize that performance could be further analyzed by testing more specialized, region-specific fine-tuning and testing. To test this, we evaluate the performance of IndicBERT V2 fine-tuned with three separate versions of the state specific dialects dataset and compare results with combined model trained on data of all 11 dialects. As shown in Table 4, the results give granular speculation and confirm this hypothesis for one of the three regions. The Himachal-specific model (93.46 F1), the Odisha-specific model (90.77 F1) both modestly perform better compared to the model trained on collective datset (91.5 and 89,77 F1 respectively). The Uttar Pradesh specific model (83.5 F1) seems to achieve a highest performance increase compared to combined model (71.66 F1), likely due to dialects being close to Himachal Pradesh Dialects causing confusion in combined model (see Fig. 2). The findings in Table 4 suggest that while a general Indic pre-training provides a strong foundation, tailoring the fine-tuning process to specific dialect families, region or in our case states is a key strategy for maximizing accuracy in dialect identification tasks.

### 4.2   Multiple choice questions

**Table 5.** Comparative performance of the combined model with state-specific models on the MCQ task.

| Model Type | Test Dataset | Prec. | Recall | F1 |
|---|---|---|---|---|
| All Dialects Combined Model | All Dialects | 77.26 | 76.42 | 77.54 |
| | Himachal Pradesh | 73.56 | 73.82 | 73.64 |
| | Uttar Pradesh | 81.54 | 80.07 | 80.77 |
| | Odisha | 79.08 | 80.56 | 79.07 |
| State Specific (Himachal) | Himachal Pradesh | 80.56 | 76.82 | 78.64 |
| State Specific (Uttar Pradesh) | Uttar Pradesh | 84.54 | 81.07 | 82.77 |
| State Specific (Odisha) | Odisha | 82.08 | 82.56 | 80.07 |

Building on the finding of the state-specific fine-tuning from the previous task of dialect classification, we extended this approach to our second NLU task: Multiple-Choice Question Answering (MCQ). We again fine-tune IndicBERT V2 models on combined 11 dialects data and state specific subsets and evaluate them on specific test dataset from different dialect families. The results are consistent with our earlier findings, showing that the specialized model subset significantly outperformed the baseline collective model that had been trained and tested on all dialects simultaneously. The results, presented in Table 5, demonstrate strong performance across all three dialect subsets, with F1-scores of 73.64 (Himachal Pradesh), 80.77 (Uttar Pradesh), and 79.07 (Odisha) raised to 78.64, 82.77 and 80.07 respectively via state-specific models. This reinforces that fine-tuning on the INDIC-DIALECT dataset enables a strong pre-trained model like IndicBERT V2 to handle complex NLU tasks beyond simple classification. Also, it is important to note that MCQ task is more challenging (have comparatively lower F-scores) than the classification task due to strong distractors used in different options of the MCQ dataset, and its connection to the translations of question in different options. With this, we are now set to evaluate the performance of machine translation (MT) task on the proposed INDIC-DIALCT dataset, which we discuss in the subsequent paragraphs.

### 4.3   Dialect Translation

To rigorously evaluate our translation models, we first established strong, direction-specific baselines. Based on literature large-scale low-resource MT efforts (e.g., NLLB) highlight the need for targeted resources at the variety level (NLLB Team 2022). We fine-tuned two separate IndicBERTSS (a generative variant of IndicBERT V2 required for translation task) models: one exclusively for the dialect-to-language task, and a second exclusively for the language-to-dialect translation task. The specialized AI models are then used for a comparison with our proposed approaches. As we will see in the subsequent sections, for the dialect-to-language translation, the hybrid AI model which uses a dictionary generated sentence in conjection with original source sentence cause substantial performance increase over its corresponding fine-tuned baseline. For the language-to-dialect direction, the rule-based AI model which uses dictionary generated sentence as input for AI model, significantly outperform its fine-tuned baseline. The direct comparison framework validates that the specialized rule-based AI and hybrid AI architectures are more effective for the nuances of dialectal translation than a standard fine-tuning approach. We also mention results for translation using only simple bilingual dictionary as "rule-based" in the result Tables (6, 7), which we discuss in next sections.

### 4.3.1   Dialect-to-Language Translation

**Table 6.** Performance of rule-based (bilingual dictionary based) translation and different AI based translation approaches for dialect-to-language translation.

| Model | BLEU Scores % |
|---|---|
| Rule-based | 23.36 |
| AI model | 46.57 |
| Rule-based followed by AI | 54.21 |
| Hybrid AI Model | **61.32** |

For the dialect-to-language translation, we evaluate different approaches with fine-tuned IndicBERTSS as AI model. The hybrid AI approach yield the highest performance of the translation experiments, see Table 6, achieving a BLEU score of 61.32% and substantially outperforming the baseline's score of 23.36%. The result highlights the complementary strengths of AI model + rule-based systems. The AI component is essential for interpreting the greater lexical and syntactic variability inherent in the dialectal input, while the rule-based processing layer refines the output by enforcing grammatical consistency and standardizing terminology. This combination proved most effective for mapping diverse, low-resource inputs to a high-resource, standardized output.

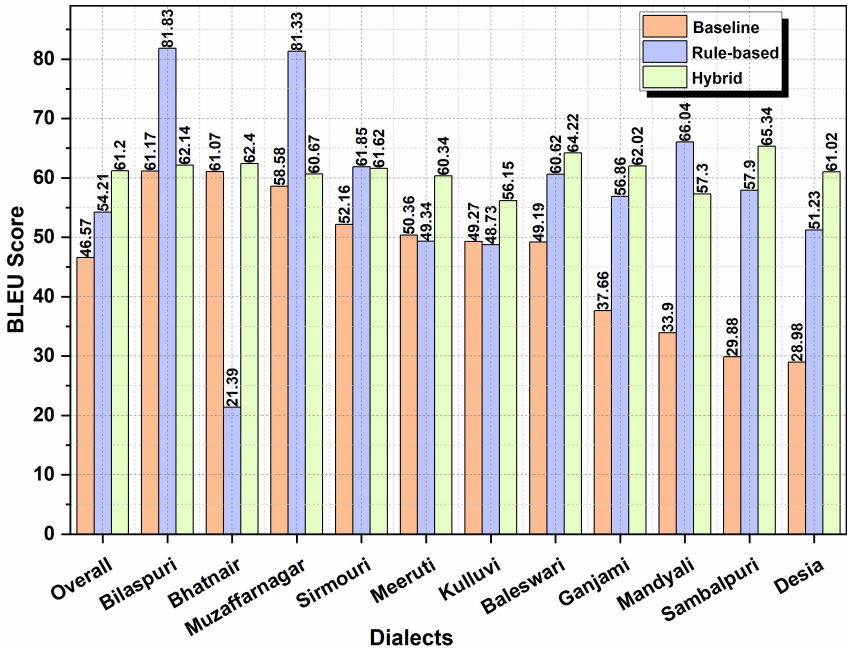

**Figure 5.** Dialect specific details of dialect-to-language translation: BLEU score comparison for different AI based approaches (rows 2-4 from Table 6).

### 4.3.2   Language-to-Dialect Translation

For the standard language-to-dialect translation task, we benchmarked our proposed Hybrid Model against a dedicated IndicBERTSS baseline that was fine-tuned exclusively for this direction, see Table 7. The results demonstrated the clear superiority of the rule based

## Dialect to language Translation

| | |
|---|---|
| **source sentence Ganjami:** | ଗୀ ସାଙ୍ଗ ଗୋଟେ ଇ-କମର୍ସ ଷ୍ଟୋରଟେ ଆରମ୍ଭ କରିଛି। |
| **Model Generated Hindi:** | मेरी सहेली ने ई-कॉमर्स स्टोर शुरू किया है। |
| **Correct Sentence:** | मेरी सहेली ने ई-कॉमर्स स्टोर शुरू किया है। |
| **source sentence Meeruti:** | हमारे यहाँ शादियाँ बड़े जोरसोर से होवे हैं। |
| **Model Generated Hindi:** | हमारे यहाँ शादियाँ बड़े धूमधाम से होती हैं। |
| **Correct Sentence:** | हमारे यहाँ शादियाँ बड़े धूमधाम से होती हैं। |
| **source sentence Bhatnair:** | मेरे भाई गणित ओलंपियाड मै राष्ट्रीय स्तर पै जीतौ। |
| **Model Generated Hindi:** | मेरे भाई ने गणित ओलंपियाड में राष्ट्रीय स्तर पर जीत हासिल की। |
| **Correct Sentence:** | मेरे भाई ने गणित ओलंपियाड में राष्ट्रीय स्तर पर जीत हासिल की। |
| **source sentence Kulluvi:** | स्वस्थ जीवन कथे सा रोज व्यायाम हो धोरा खाना जरूरी। |
| **Model Generated Hindi:** | स्वस्थ जीवन के लिए हर जरूरी व्यायाम और  आहार जरूरी है। |
| **Correct Sentence:** | स्वस्थ जीवन के लिए नियमित व्यायाम और संतुलित आहार जरूरी है। |

**Figure 6.** Machine Translation predictions examples using the best performing Model for translating dialect into standard language.

**Table 7.** Performance of rule-based (bilingual dictionary based) translation and different AI based translation approaches for language-to-dialect translation..

| Model | BLEU Scores % |
|---|---|
| Rule-based | 31.71 |
| AI Model | 27.59 |
| Rule-based followed by AI | **48.44** |
| Hybrid AI Model | 33.82 |

approach, which achieves a BLEU score (Papineni et al. 2002) of 48.44% a significant improvement of over 20 points compared to the fine-tuned baseline's score of 27.59%. The outcome validates our hypothesis that the translation from a standardized, high-resource language to its regional variant is often a process of applying systematic and predictable transformations. These transformations, including consistent phonetic shifts and lexical substitutions, can be more effectively captured by a well-defined rule set than by a purely data-driven AI model in a low-resource setting.

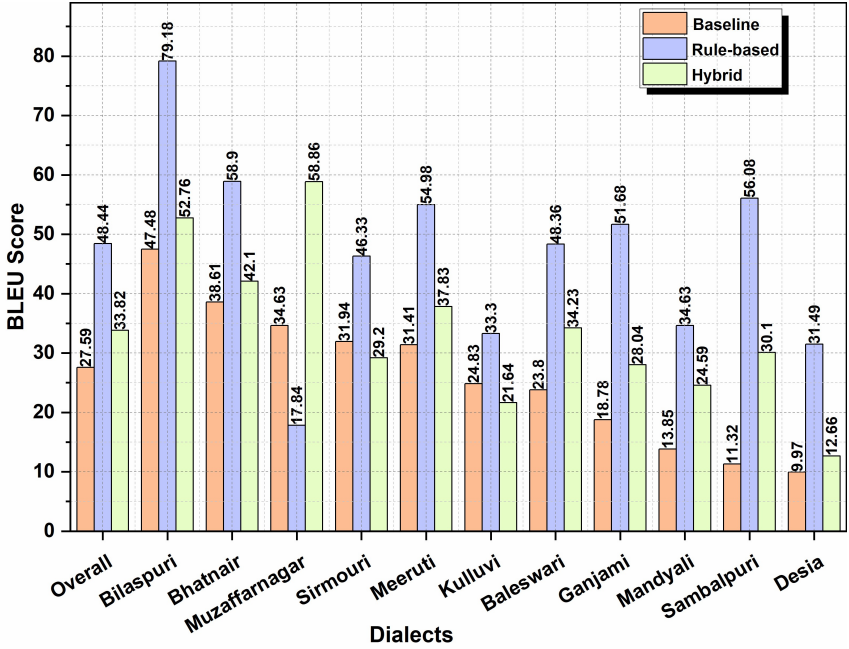

**Figure 7.** Dialect specific details of language-to-dailect translation: BLEU score comparison for different AI based approaches (rows 2-4 from Table 7).

## language to dialect Translation

| | |
|---|---|
| **source sentence :** | हमारे गाँव के बच्चों ने तालाब की सफाई की। |
| **Model Generated Mandyali:** | आसारे गाँव रे बच्चे तालाब री सफाई किती। |
| **Corrcet Sentence:** | हमारे गाँव के बच्चों ने तालाब की सफाई की। |
| **source sentence:** | मेरे पापा ने मैराथन दौड़ में भाग लिया और अच्छा प्रदर्शन किया। |
| **Model Generated Sirmouri:** | मेरे पापा रे मैराथन दौड़ दी भाग लोई और अच्छा प्रदर्शन कोरा असो |
| **Corrcet Sentence:** | मेरे पापा ने मैराथन दौड़ दा भाग लिया साथी अच्छा प्रदर्शन कोरा। |
| **source sentence:** | मेरी मम्मी को योगा ट्रेनर ने व्यायाम सिखाया। |
| **Model Generated Sambalpuri:** | ମୋର ମାଆ ବରକୁ ଗୁଚ୍ଚ ଯୋଗ ପ୍ରଶିକ୍ଷକ ବ୍ୟାୟାମ ଶିଖେଇଥିଲେ। |
| **Corrcet Sentence:** | ମୋର ମାଆଙ୍କେ ଯନେ ଯୋଗ ପ୍ରଧିକ୍ଷକ ବ୍ୟାୟାମ୍ ଧିଖେଇଥିଲେ। |
| **source sentence:** | दीपा ने जैविक खेती पर एक वर्कशॉप आयोजित की। |
| **Model Generated Desia:** | ଦୀପା ଜୈବିକ ଚାଷ୍ ଉପ୍ରେ ଗୋଟେ କର୍ମଘାଲା ପରିଚାଲନା କରିଦେଲ। |
| **Corrcet Sentence:** | ଦୀପା ଜୈବିକ ଚାଷ୍ ଉପ୍ରେ ଗଟେକ୍ କର୍ଶାଳା ପରିଚାଲନା କରି ଦେଲା।। |

**Figure 8.** Machine Translation predictions examples using the best performing Model for translating language into dialect.

### 4.4  Analysis of Dialectal Proximity: Embedding Space and Lexical Overlap

The performance disparity observed between high accuracy on the Dialect Classification task and the mixed results on the MCQ and MT tasks is rooted in the close linguistic and representational proximity of the low-resource dialects to their standard parent languages (Hindi and Odia). This section analyzes this underlying challenge using the internal geometry of the IndicBERT embedding space and the calculated lexical distance.

### 4.4.1 IndicBERT V2 Embedding Space Analysis

We analyzed the sentence embeddings (specifically the `[CLS]` token) extracted from the fine-tuned IndicBERT model. When these high-dimensional vectors are projected into a 2D space, via t-SNE (Maaten and Hinton 2008), they reveal clear, but highly overlapping, clustering behavior:

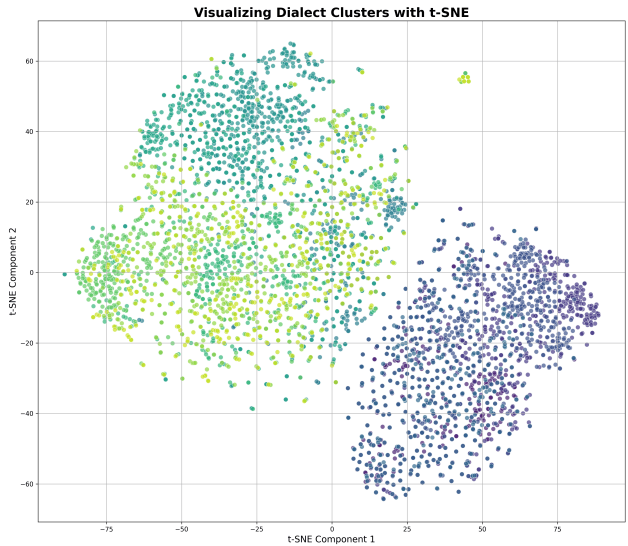

**Figure 9.** The plot shows the clustering of the [CLS] token representations, demonstrating the clear separation between the two language families (Hindi and Odia). Critically, the tight, overlapping proximity of the standard language (center of each cluster) with its regional dialects confirms the high linguistic similarity, which contributes to model confusion in fine-grained NLU and MT tasks.

- Clustering by Family: The embeddings strongly cluster into two distinct, high-density regions corresponding to the Hindi Dialect Family and the Odia Dialect Family. This confirms that the model successfully learns the broad linguistic differences between the two parent languages.
- Intra-Family Overlap and Confusion: Crucially, within each family cluster, the standard language (Hindi or Odia) and its constituent dialects lie in extremely close proximity. For instance, the representations of Kulluvi, Bilaspuri, and Mandyali overlap significantly with standard Hindi. This spatial closeness in the embedding space indicates that the pre-trained knowledge base of IndicBERT views these dialects as highly similar variants of the standard language.
- Implication: This tight overlap suggests that the model's high classification accuracy is achieved by learning a fragile, non-linear decision boundary to separate closely related data points, which explains the difficulty in the more sensitive tasks (MCQ, MT) where subtle lexical and morphological differences are essential. Errors in MCQ and MT are a direct consequence of the model's propensity to confuse these closely correlated representations.

### 4.4.2   Lexical Proximity via Edit Distance

To provide a non-model-based, quantitative measure of linguistic closeness, we computed the average character-level Levenshtein (Levenshtein 1966) Edit Distance between the dialectal sentences and Hindi language translations across the parallel corpus. This metric quantifies the minimum number of single-character operations required for transformation, thus serving as a proxy for orthographic and phonetic difference.

As shown conceptually in Figure 2, the average distance for all 11 dialects except odia based from Hindi language is consistently low.

- Confirmation of Similarity: The low edit distance validates the observation from the embedding analysis: these are dialects, not distinct languages, and they share significant lexical and morphological overlap with their parent.
- Correlation with Model Performance: The dialects exhibiting the lowest edit distances (i.e., those most lexically similar to the standard Hindi language, such as the Meerut dialect near Hindi) were found to be the most challenging for the fine-tuned MT models, particularly when generating the unique dialectal forms. This correlation suggests that the model's struggle is inherent to the high inter-class similarity of the data points, which the model's encoder fails to fully disentangle based on distributional features alone.

This dual analysis—both in the abstract embedding space and the concrete lexical space— underscores why specialized, Translation strategies (as detailed in Section 4) were necessary to overcome the inherent ambiguity of these closely related linguistic varieties.

## 5.   Conclusion

In this work, we introduce INDIC-DIALECT, a novel parallel corpus for eleven Hindi and Odia dialects, validated with a high inter-annotator agreement. We prove our dataset's efficacy: while zero-shot LLMs fail on our tasks, models fine-tuned on INDIC-DIALECT achieve substantial performance gains and high accuracy.

Our analysis reveals a critical challenge: high linguistic proximity, as shown in our lexical distance analysis, creates task-dependent ambiguity. For example, the UP dialects, being lexically very close to Hindi, performed worst in classification due to confusion, but best in MCQ as the task benefited from the small linguistic gap. However, this proximity proved detrimental in translation, where model confusion suppressed dialectal nuance, resulting in poor baseline performance.

These finding validate our INDIC-DIALECT benchmark efficacy:. We Observed that a Rule-Based Model is superior for the standard language-to-dialect direction by explicitly enforcing nuance, while a Hybrid Model is optimal for the dialect-to-standard task. We establish that Indic-specific models are superior backbones and that specialized architectures are beneficial for dialectal NLP. Along with our learning we are releasing INDIC-DIALECT as a resource to foster research into more equitable, linguistically-aware models for indic NLP.

## 6.   Limitations

While INDIC-DIALECT is set to publish dataset as open source our methodology for Indian dialect NLP, we acknowledge several limitations that present opportunities for future research. First, our study is focused on a specific set of dialects from the Hindi and Odia families within the Indo-Aryan language group; the findings and models may not generalize to the vast linguistic diversity of other language families in India, such as Dravidian or Tibeto-Burman. Second, as our parallel corpus was created by translating standard language sentences, it may not fully capture the organic, dialectal speech, which often includes extensive code-switching. Addressing these limitations will be a crucial next step toward building a truly comprehensive and robust ecosystem for Indian dialect.

## References

Abdul-Mageed, Muhammad, Chiyu Zhang, Houda Bouamor, and Nizar Habash. 2020. NADI 2020: the first nuanced Arabic dialect identification shared task. In Proceedings of the fifth arabic natural language processing workshop, edited by Imed Zitouni, Muhammad Abdul-Mageed, Houda Bouamor, Fethi Bougares, Mahmoud El-Hajj, Nadi Tomeh, and Wajdi Zaghouani, 97–110. Barcelona, Spain (Online): Association for Computational Linguistics, December. https://aclanthology.org/2020.wanlp-1.9/.

Bafna, Niyati. 2022. Empirical models for an indic language continuum.

Bandarkar, Lucas, Davis Liang, Benjamin Muller, Mikel Artetxe, Satya Narayan Shukla, Donald Husa, Naman Goyal, Abhinandan Krishnan, Luke Zettlemoyer, and Madian Khabsa. 2023. The belebele benchmark: a parallel reading comprehension dataset in 122 language variants. arXiv preprint arXiv:2308.16884.

Bouamor, Houda, Sabit Hassan, and Nizar Habash. 2019. The MADAR shared task on Arabic fine-grained dialect identification. In Proceedings of the fourth arabic natural language processing workshop, edited by Wassim El-Hajj, Lamia Hadrich Belguith, Fethi Bougares, Walid Magdy, Imed Zitouni, Nadi Tomeh, Mahmoud El-Haj, and Wajdi Zaghouani, 199–207. Florence, Italy: Association for Computational Linguistics, August. https://doi.org/10.18653/v1/W19-4622. https://aclanthology.org/W19-4622/.

Census of India 2011: Language Data. https://censusindia.gov.in/nada/index.php/catalog/42561. Office of the Registrar General  Census Commissioner, India.

Choudhary, Narayan. 2021. Ldc-il: the indian repository of resources for language technology. Language Resources and Evaluation 55:855–867. https://doi.org/10.1007/s10579-020-09523-3. https://link.springer.com/article/10.1007/s10579-020-09523-3.

Cohen, Jacob. 1960. A coefficient of agreement for nominal scales. Educational and Psychological Measurement 20 (1): 37–46. https://journals.sagepub.com/doi/10.1177/001316446002000104.

Doddapaneni, Sumanth, Rahul Aralikatte, Gowtham Ramesh, Shreya Goyal, Mitesh M. Khapra, Anoop Kunchukuttan, and Pratyush Kumar. 2023. Towards leaving no Indic language behind: building monolingual corpora, benchmark and models for Indic languages. In Proceedings of the 61st annual meeting of the association for computational linguistics (volume 1: long papers), 12402–12426. Toronto, Canada: Association for Computational Linguistics, July. https://doi.org/10.18653/v1/2023.acl-long.693. https://aclanthology.org/2023.acl-long.693/.

Gala, Jay, Pranjal Behl, et al. 2023. Indiccorp v2: towards building inclusive and actionable corpora for indian languages. In Proceedings of the 2023 conference on empirical methods in natural language processing, 3059–3080.

Hu, Junjie, Sebastian Ruder, Aditya Siddhant, Graham Neubig, Orhan Firat, and Melvin Johnson. 2020. XTREME: a massively multilingual multi-task benchmark for evaluating cross-lingual generalisation. In Proceedings of the 37th international conference on machine learning. Proceedings of Machine Learning Research. PMLR. https://proceedings.mlr.press/v119/hu20b.html.

Indian Institute of Science, ARTPARK, and Google. 2022. Project vaani: scaling the digital divide. https://vaani.iisc.ac.in. Accessed: 2025-12-23.

Javed, Tahir, Janki Nawale, Sakshi Joshi, Eldho George, Kaushal Bhogale, Deovrat Mehendale, and Mitesh M. Khapra. 2024. Lahaja: a robust multi-accent benchmark for evaluating hindi asr systems. In Interspeech 2024. Kos, Greece: ISCA. https://www.isca-archive.org/interspeech_2024/javed24_interspeech.pdf.

Joshi, Pratik, Sebastin Santy, Amar Budhiraja, Kalika Bali, and Monojit Choudhury. 2020. The state and fate of linguistic diversity and inclusion in the nlp community. In Proceedings of the 58th annual meeting of the association for computational linguistics, 6282–6293. https://aclanthology.org/2020.acl-main.560/.

Khanuja, Simran, Sandipan Dandapat, Anirudh Srinivasan, Sunayana Sitaram, and Monojit Choudhury. 2021. Muril: multilingual representations for indian languages. In Proceedings of the 2021 conference of the north american chapter of the association for computational linguistics: human language technologies, 3449–3456.

Kreutzer, Julia, Isaac Caswell, Lisa Wang, Ahsan Wahab, Daan van Esch, Nasanbayar Ulzii-Orshikh, Allahsera Tapo, et al. 2022. Quality at a glance: an audit of web-crawled multilingual datasets. Edited by Brian Roark and Ani Nenkova. Transactions of the Association for Computational Linguistics (Cambridge, MA) 10:50–72. https://doi.org/10.1162/tacl_a_00447. https://aclanthology.org/2022.tacl-1.4/.

Kumar, Aman, Himani Shrotriya, Prachi Sahu, Raj Dabre, Ratish Puduppully, Anoop Kunchukuttan, Amogh Mishra, Mitesh M. Khapra, and Pratyush Kumar. 2022. Indicnlg benchmark: multilingual datasets for diverse NLG tasks in indic languages. In Proceedings of the 2022 conference on empirical methods in natural language processing. Abu Dhabi, United Arab Emirates: Association for Computational Linguistics. https://aclanthology.org/2022.emnlp-main.360/.

Levenshtein, Vladimir I. 1966. Binary codes capable of correcting deletions, insertions, and reversals. Soviet Physics Doklady 10:707–710. http://mi.mathnet.ru/eng/dan24228.

Lewis, Mike, Yinhan Liu, Naman Goyal, et al. 2020. Bart: denoising sequence-to-sequence pre-training for natural language generation, translation, and comprehension. In Proceedings of the 58th annual meeting of the association for computational linguistics, 7871–7880.

Liu, Yinhan, Jiatao Gu, Naman Goyal, Xian Li, Sergey Edunov, et al. 2020. Multilingual denoising pre-training for neural machine translation. Transactions of the Association for Computational Linguistics 8:726–742.

Maaten, Laurens van der, and Geoffrey Hinton. 2008. Visualizing data using t-SNE. Journal of Machine Learning Research 9:2579–2605. http://www.jmlr.org/papers/v9/vandermaaten08a.html.

Mhaske, Arnav, Harshit Kedia, Sumanth Doddapaneni, Mitesh M Khapra, Pratyush Kumar, Rudra Murthy V, and Anoop Kunchukuttan. 2022. Naamapadam: a large-scale named entity annotated data for indic languages. arXiv preprint arXiv:2212.10168.

Nekoto, Wilhelmina, Julia Kreutzer, Vukosi Marivate, et al. 2020. Participatory research for low-resourced machine translation: a case study in african languages. In Findings of the association for computational linguistics: emnlp 2020, 2144–2160.

NLLB Team. 2022. No language left behind: scaling human-centered machine translation. arXiv: 2207.04672 [cs.CL]. https://arxiv.org/abs/2207.04672.

OpenAI. 2023. Gpt-4 technical report. arXiv preprint arXiv:2303.08774, https://arxiv.org/abs/2303.08774.

Papineni, Kishore, Salim Roukos, Todd Ward, and Wei-Jing Zhu. 2002. Bleu: a method for automatic evaluation of machine translation. In Proceedings of the 40th annual meeting of the association for computational linguistics, 311–318. Philadelphia, Pennsylvania, USA. https://aclanthology.org/P02-1040/.

Reid, Machel, Nikolay Savinov, Denis Teplyashin, et al. 2024. Gemini 1.5: unlocking multimodal understanding across millions of tokens of context. arXiv preprint arXiv:2403.05530.

Samin, Md Nazmus Sadat, Jawad Ibn Ahad, Tanjila Ahmed Medha, Fuad Rahman, Mohammad Ruhul Amin, Nabeel Mohammed, and Shafin Rahman. 2024. Bangladialecto: an end-to-end ai-powered regional speech standardization. In 2024 ieee international conference on big data (bigdata), 1635–1644. IEEE.

Sanh, Victor, Lysandre Debut, Julien Chaumond, and Thomas Wolf. 2019. Distilbert, a distilled version of bert: smaller, faster, cheaper and lighter. arXiv preprint arXiv:1910.01108.

Scherrer, Yves, and Owen Rambow. 2010. Word-based dialect identification with georeferenced rules. In Proceedings of the 2010 conference on empirical methods in natural language processing, edited by Hang Li and Lluís Màrquez, 1151–1161. Cambridge, MA: Association for Computational Linguistics, October. https://aclanthology.org/D10-1112/.

Xu, Fan, Mingwen Wang, and Maoxi Li. 2017. Sentence-level dialects identification in the greater china region. arXiv preprint arXiv:1701.01908.

Zampieri, Marcos, Preslav Nakov, Nikola Ljubešić, Jörg Tiedemann, Shervin Malmasi, and Ahmed Ali. 2018. Language identification and morphosyntactic tagging: the second vardial evaluation campaign. In Proceedings of the fifth workshop on nlp for similar languages, varieties and dialects (vardial 2018), 1–17. Santa Fe, New Mexico, USA: Association for Computational Linguistics. https://aclanthology.org/W18-3901/.

Zellers, Rowan, Ari Holtzman, Yonatan Bisk, Ali Farhadi, and Yejin Choi. 2019. Hellaswag: can a machine really finish your sentence? arXiv preprint arXiv:1905.07830.